\documentclass[conference]{IEEEtran}
\IEEEoverridecommandlockouts
\usepackage{cite}
\usepackage{amsmath,amssymb,amsfonts}
\usepackage{algorithmic}
\usepackage{graphicx}
\usepackage{textcomp}
\usepackage{xcolor}

\usepackage{microtype}
\usepackage{graphicx}
\usepackage{subfigure}
\usepackage{booktabs}
\usepackage[bookmarks=false]{hyperref}

\usepackage{times}
\usepackage{helvet}
\usepackage{courier}
\usepackage{amsmath,amssymb}
\usepackage{mathrsfs}
\usepackage{booktabs}
\usepackage{color}
\usepackage{courier}
\usepackage{epsfig}
\usepackage{graphicx}
\usepackage{subfigure}
\usepackage{helvet}
\usepackage{multirow}
\usepackage{times}
\usepackage{url}
\usepackage{enumitem}

\newcommand\etal{{et~al.}}
\newcommand\ie{{i.e.}}

\newcommand\eg{{e.g.}}

\def\BibTeX{{\rm B\kern-.05em{\sc i\kern-.025em b}\kern-.08em
    T\kern-.1667em\lower.7ex\hbox{E}\kern-.125emX}}
\begin{document}

\title{Fast Automatic Feature Selection for Multi-Period Sliding Window Aggregate in Time Series\\
}

\author{\IEEEauthorblockN{1\textsuperscript{st} Rui An}
\IEEEauthorblockA{\textit{Data Lab} \\
\textit{ZhongAn Technology}\\
Shanghai, China \\
anrui001@zhongan.com
}
\and
\IEEEauthorblockN{2\textsuperscript{nd} Xingtian Shi}
\IEEEauthorblockA{\textit{Data Lab} \\
\textit{ZhongAn Technology}\\
Shanghai, China \\
shixingtian@zhongan.com
}
\and
\IEEEauthorblockN{3\textsuperscript{rd} Baohan Xu}
\IEEEauthorblockA{\textit{Bilibili} \\
\textit{Bilibili}\\
Shanghai, China \\
xubaohan@bilibili.com
}
}

\maketitle

\begin{abstract}
    As one of the most well-known artificial feature sampler,
    the sliding window is widely used in scenarios where spatial and temporal information exists,
    such as computer vision, natural language process, data stream, and time series.
    Among which time series is common in many scenarios like credit card payment, user behavior, and sensors.
    General feature selection for features extracted by sliding window aggregate calls for time-consuming iteration to generate features,
    and then traditional feature selection methods are employed to rank them.
    The decision of key parameter, \ie~the period of sliding windows, depends on the domain knowledge and calls for trivial.
    Currently, there is no automatic method to handle the sliding window aggregate features selection.
    As the time consumption of feature generation with different periods and sliding windows is huge,
    it is very hard to enumerate them all and then select them.
    
    In this paper, we propose a general framework using Markov Chain to solve this problem.
    This framework is very efficient and has high accuracy,
    such that it is able to perform feature selection on a variety of features and period options.
    We show the detail by 2 common sliding windows and 3 types of aggregation operators.
    And it is easy to extend more sliding windows and aggregation operators in this framework by employing existing theory about Markov Chain.
\end{abstract}

\begin{IEEEkeywords}
    AutoML, Sliding Window Aggregate, Feature Selection, Time Series
\end{IEEEkeywords}

\section{Introduction}

The time series scenario can be simplified into an entity table $T_e$ with an action table $T_a$, while the foreign key of $T_a$ is the primary key of $T_e$,
and columns in $T_a$ are time series of the action of each entity.

Sliding window aggregate operators scan the action records in the time dimension and then aggregate them to extract some periodic features.
This process by given sliding window aggregator on a given feature in $T_a$ of a given entity can be summarized as follows:
\begin{enumerate}[leftmargin=1.2cm, label=Step \arabic*:]
    \item Resample a new column, denoted as $\mathcal{C}$, from the given column of all the action records of a given entity in $T_a$ by a given sample frequency.
    Denote the count of timestamps in $\mathcal{C}$ after resampling as $\ell$.
    \item Use a sliding window $W$ with period $w$ to scan $\mathcal{C}$ and calculate the corresponding value in each timestamp,
    and construct a new column.
    \item Use aggregation operator $AGG$ to extract a value over the constructed column.
\end{enumerate}

\begin{figure}
\centering
\includegraphics[width=\linewidth]{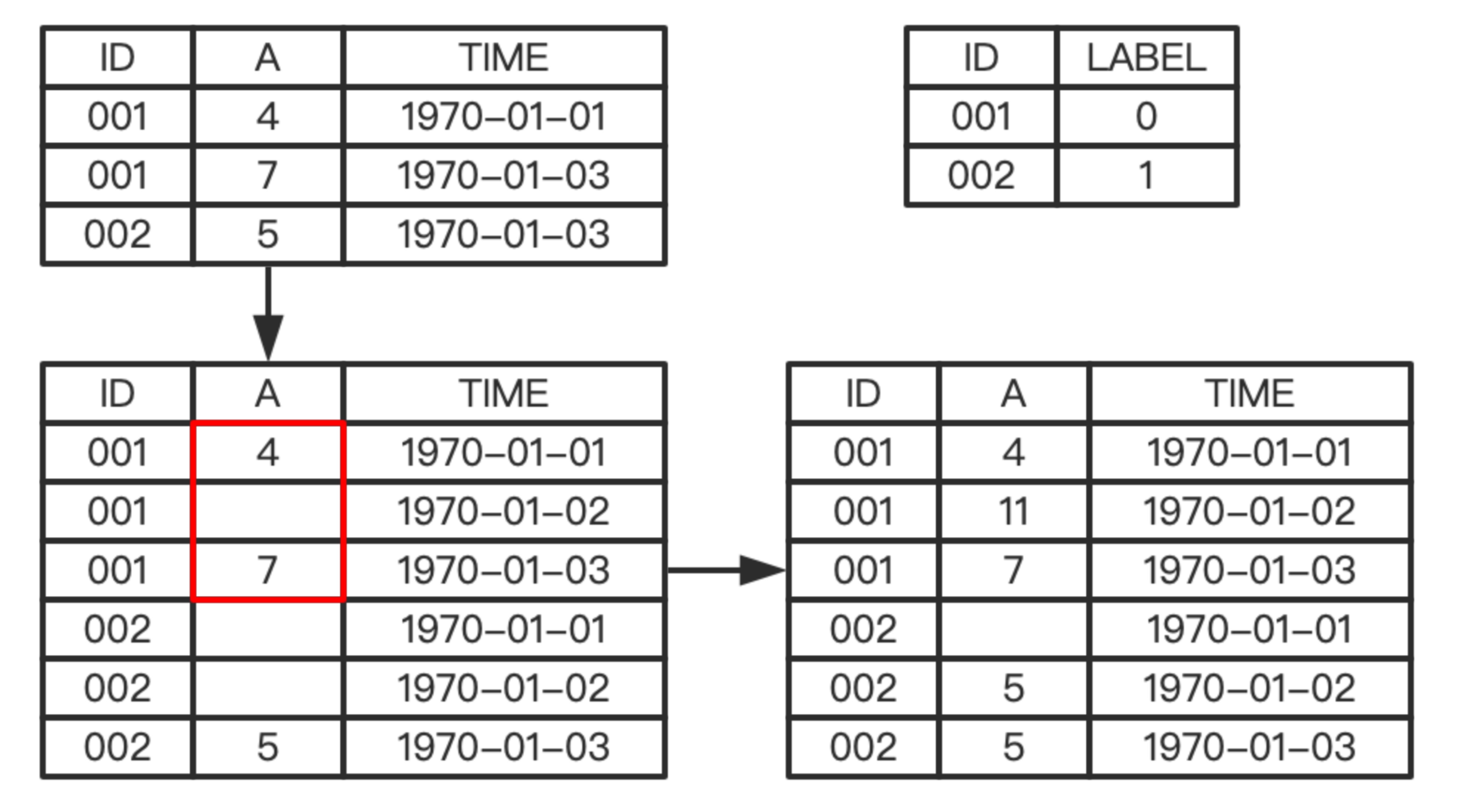}
\caption{
    Example of sum sliding window with period 3 days.
    The left-top table is $T_a$ with foreign key ID and feature column A,
    while the right-top table is $T_e$ whose primary key is ID.
    The left-bottom table is resampled with frequency 1 day from $T_a$.
    The sliding window scans feature A by TIME column for each entity.
    When it scans to timestamp 1970-01-02 of entity 001, all records in recent 3 periods,~\ie~the red box, are summed.
    Then the intermediate table (right bottom) is used for aggregate to extract feature of each entity.
}
\label{fig:sliding_window}
\end{figure}

The extracted value is a feature of the given entity,
and the values extracted from the given column of all entities will then form a feature column,
while all feature columns extracted by different periods, sliding windows and aggregate operators will then form a feature table $T_f$.
$T_f$ and the label column in $T_e$ are used to train a model.
Intuitively, Fig.~\ref{fig:sliding_window} shows the process of a sum sliding window with period 3 days.
We abuse the below equation to represent this process:
\begin{equation}
    T_f = AGG(W(T_a, w))
\end{equation}

Sliding window aggregation is widely used in real-world applications for its good interpretability,
for example, max aggregate on sum sliding window in the credit card scenario can measure the maximum of total expenditure for a user in a period.
In practice $T_f$ is always generated by brute-force iteration while the period w are designed with intuition or domain knowledge,
and then general method is used to select from $T_f$.
Since automatic solution must try different w, and the generation from original time series tables calls for massive iteration,
it must be quick enough to process variety of period options.

Our key thought to ameliorate this process is to contact labels with Ta directly,
which means the observation of $T_a$ but not $T_f$ is used to select features.
The key contributions are summarized as follows:
\begin{enumerate}
    \item We present a framework to optimize the automatic selection of sliding window aggregate features.
    The speed and accuracy are good enough to process massive period options and sliding window aggregators.
    Thus it can support the end-to-end AutoML solution for time series.
    \item We set up a general model of sliding window aggregator by stochastic process theory, and make detailed analysis and derivation.
    The work is helpful when extending more sliding window aggregators,
    and this probability model, as well as its estimation method will be useful for further research on similar scenarios.
\end{enumerate}

In this paper we use 2 sliding windows, \emph{sum window}, and \emph{average window},
as well as 3 aggregation operators, max value, min value and average value.
The combination of sliding windows and aggregation operators then conducts 6 sliding window aggregators.
The design of sliding window aggregators is not the main purpose of this paper,
they can be extended into this framework by similar methods.

The detail of this framework is in Section~\ref{sec:framework}
and experiments in Section~\ref{sec:exp} will show its high accuracy and speed.

\section{Related Works}
\label{sec:related}

AutoML is a hot spot in recent years, and receives some significant progress.
But most recent research focus on the hyperparameters optimization, Network Architecture Search and model selection methods in different scenarios,
automatic methods about feature extraction from the original database are few.
One Button Machine~\cite{Lam2017One} and Deep Feature Synthesis~\cite{Kanter2015Deep}
enumerate transformations and aggregations on whole related tables to extract features,
but time series is not the topic of them.

Feature selection is a well-discussed topic~\cite{Tang2014Feature},
many theories, methods, and criteria are employed to it.
Generally speaking, supervised feature selection methods~\cite{Molina2003Feature} evaluate features by the contribution to the object function,
while unsupervised feature selection methods~\cite{Li:2019:AUF:3292500.3330856} evaluate them by the ability to reconstruct feature space.
Besides, some methods focus on the elimination of the redundancy in streaming features~\cite{2017A, 2018Online}.
The streaming feature scenario may be confused with time series, but they are actually different,
the input of this scenario in time dimension is feature but not timestamp of feature.
Almost all feature selection methods only reponse to a generated feature space,
currently we have not found existing methods on the end-to-end sliding window aggregate feature selection from the original database.

The explicit feature space $T_f$ can not be known before iteration and calculation.
Hellerstein~\etal~\cite{Hellerstein1997Online} proposed some basic ideas about online aggregation,
they estimate the confidence bound of the result by the observation of original table.
Aggregation on the sliding window is more complex than direct aggregation, but can still be handled with a similar idea.

In addition, some theoretical research on stochastic process and Markov Chain help us a lot.
The aggregate operators we use in this paper are related to the average and extreme value of Markov Chain.
Works on the distribution of average value of Markov Chain, \eg~\cite{2012arXiv1201.0559C}, ~\cite{2018arXiv180611519R} and ~\cite{2018arXiv180200211F},
and the Extreme Value Theory (EVT) of stochastic process, \eg~~\cite{O1987Extreme},~\cite{Rootz1988Maxima},~\cite{Leadbetter1988Extremal},~\cite{Perfekt1994Extremal},
~\cite{Perfekt1997Extreme} and~\cite{Gikhman2006A},
solve the problem we met in the derivation of formulation.

\section{Framework}
\label{sec:framework}

The proposed framework aims at automatic selection from sliding window aggregate features with multi period setting.
In order to support the variety of period, this framework try to estimate the distribution of $T_f$ under different period setting instead of generate it.
This framework is divided into 3 steps as follows:
\begin{enumerate}[leftmargin=1.2cm, label=Step \arabic*:]
    \item Fit some basic parameters from $T_a$. We will list them in Section~\ref{sec:assumption}.
    \item Set up the stochastic process model with given parameters
    and estimate the confidence bound of aggregate value on sliding windows.
    \item Sample from the estimated bound to form a \emph{fake} $T_f$, train a model and get the feature importance.
    Repeat it many times to get average feature importance.
\end{enumerate}

The third step is an ensemble strategy, \ie~resample from distribution to get better generalization ability.
Even though we have not constructed \emph{real} $T_f$, the estimated feature importance is gotten by observing $T_a$.
In this section we describe the probability model of sliding window, the estimation of aggregate on sliding window,
as well as the complexity analysis of this method.

Recall that $\mathcal{C}$ represents the given column of the given entity in $T_a$.
In this section, we show the detail of the proposed framework on $\mathcal{C}$.
And it is easy to be promoted to whole $T_a$.

\subsection{Sliding Window Model}
\label{probmodel}

In this section we describe the probability model of both two sliding windows.

\subsubsection{Basic Assumptions}
\label{sec:assumption}

In order to describe the model mathematically, and set up a model with not that complex properties,
two assumptions are made as follows:
\begin{itemize}
    \item The count of records in each timestamp of $\mathcal{C}$ is a sequence notated by $\{A_i\}_{1\leq i\leq \ell}$ with i.i.d samples.
    \item The value of records in $\mathcal{C}$ is a sequence notated by $\{B_i\}_{1\leq i\leq \ell}$ with i.i.d samples.
\end{itemize}

The basic distribution assumption for $A_i$ is \emph{Binomial} assumption.
In the situation that more than one record can exist in a timestamp,
i.i.d Poisson distribution is used to model $A_i$, we call it \emph{Poisson} assumption.
And another common situation is that there is always one record in each timestamp,
which is called \emph{Always} assumption, and can be treated as a special case of \emph{Binomial} assumption.
The selection of assumption can be decided from $\mathcal{C}$ or set manually.

We model $A_i$ by $P(A_i) \propto \mathcal{B}(1, p)$ in \emph{Binomial} assumption,
and $P(A_i) \propto \mathcal{P}(1, p)$ in \emph{Poisson} assumption.
As we don't have a priori about the distribution of values in $\mathcal{C}$,
the value sequence $B_i$ is modeled by Gaussian distribution $P(B_i) \propto \mathcal{N}(\mu, \sigma^2)$.

With the above assumptions, $W(\mathcal{C}, w)$ is described as stochastic process.
And $AGG(W(C, w))$ can be treated as the estimation of a given stochastic process.

The min value $\underline{\mathcal{C}}$ and max value $\overline{\mathcal{C}}$,
mean value $\mu$, standard error $\sigma$, existence probability $p$ are fitted from $\mathcal{C}$ as the first step.
They are the key parameters of those distribution assumptions.

\subsubsection{Distribution}

\begin{figure}
\centering
\includegraphics[width=\linewidth]{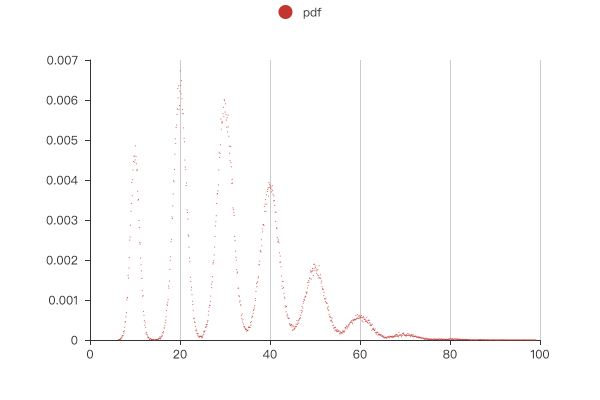}
\caption{
    PDF of 500000-length Monte Carlo simulation sum sliding windows with \emph{Binomial} assumption, $\mu=10$, $\sigma=1$, $w=10$ and $p=0.3$.
    Observations when $S(W_{i}) = 0$ are droped.
}
\label{fig:simu}
\end{figure}

The \emph{sum window} and \emph{average window} with period $w$ can be modeled as
\begin{equation}
\label{sum_window}
    W_{i}^{sum} = \sum_{n=0}^{w-1} A_{i+n} \cdot B_{i+n}
\end{equation}
\begin{equation}
\label{avg_window}
    W_{i}^{avg} = \frac{\sum_{n=0}^{w-1} A_{i+n} \cdot B_{i+n}}{\sum_{n=0}^{w-1} A_{i+n}}
\end{equation}
Denote the count of records in a window frame as
\begin{equation}
    S(W_{i}) = \sum_{n=0}^{w-1} A_{i+n}
\end{equation}

Let $m$ denotes the max appearance count of records that can happen in each timestamp, \ie~$S(W_{i}) \in [0, mw]$.

Take \emph{Binomial} assumption as an example.
Obviously $m = 1$ in \emph{Binomial} assumption.
Since ${A_i}$ is i.i.d, the distribution $P(S(W_{i})) \propto \mathcal{B}(w, p)$.
$S(W_{i})$ can also be treated as a first-order Markov Chain and the same conclusion can be gotten.

$S(W_{i})$ records occur in the observation range in timestamp $i$,
if $S(W_{i}) \neq 0$,
the distribution of the sum of $S(W_{i})$ i.i.d samples from $\mathcal{N}(\mu, \sigma^2)$ is $\mathcal{N}(S(W_{i})\mu, S(W_{i})\sigma^2)$,
and the average is $\mathcal{N}(\mu, \frac{\sigma^2}{S(W_{i})})$ since ${B_i}$ is also i.i.d.
Thus the probability models of \emph{sum window} and \emph{average window} with \emph{Binomial} assumption are
\begin{equation}
    P(W_{i}^{sum}) \propto \sum_{n=1}^w \mathcal{B}(w, p) \cdot \mathcal{N}(n\mu, n\sigma^2)
\end{equation}
\begin{equation}
    P(W_{i}^{avg}) \propto \sum_{n=1}^w \mathcal{B}(w, p) \cdot \mathcal{N}(\mu, \frac{\sigma^2}{n})
\end{equation}
If $S(W_{i}) = 0$, the sum and average are both $\mathcal{N}(0, 0)$.
Similar result can be gotten for other assumptions.
Fig.~\ref{fig:simu} is the PDF of Monte Carlo simulation with \emph{Binomial} assumption.

In general, the sliding window model with our basic assumptions is a stationary positive Harris Markov Chain,
whose stationary distribution is Gaussian Mixture Model~\cite{Kevin2012Mlapp} with $mw$ components.

\subsubsection{Transition Kernel}
\label{kernel}

As the assumptions about $\{A_i\}$ are all discrete distributions,
we can separate the transition kernels into different situations to make analysis easier.

Consider \emph{sum window} in \emph{Binomial} assumption as the example, suppose $S(W_{i}) = n$,
which means $W_{i}$ is in the $n$-th Gaussian component of GMM,
the distribution of the $n$-th Gaussian component is $\mathcal{N}(n\mu, n\sigma^2)$.

Since the next state $W_{i + 1}$ contains records from the $(i + 1)$-th to $(i + w)$-th timestamps.
the difference between $W_{i}$ and $W_{i + 1}$ only depends on the first item of current state and the newly-incoming item,
\ie~the $i$-th and the $(i + w)$-th records.
In this situation, $S(W_{i+1})$ can only be $n-1$, $n$ and $n+1$.
The transition kernel of this situation can be separated into four components as follows:

\begin{enumerate}
    \item both the first item of current state and the newly-incoming item exist,
    \ie~$A_i = 1$ and $A_{i+w} = 1$, then $S(W_{i+1}) = n$.
    \item the first item of current state exists and the newly-incoming item does not exist,
    \ie~$A_i = 1$ and $A_{i+w} = 0$, then $S(W_{i+1}) = n - 1$.
    \item the first item of current state does not exist and the newly-incoming item exists,
    \ie~$A_i = 0$ and $A_{i+w} = 1$, then $S(W_{i+1}) = n + 1$.
    \item both the first item of current state and the newly-incoming item do not exist,
    \ie~$A_i = 0$ and $A_{i+w} = 0$, this situation is special, the state will not transfer.
\end{enumerate}

Abuse $\mathcal{B}_w(x)$ to represent the probability of $v = x$ for a random variable of $v \propto \mathcal{B}(w, p)$ (where $p$ is already given).
The probability $A_i = 1$ given $S(W_{i}) = n$ is
\begin{align}
    P(A_i = 1|S(W_{i}) = n) &= \frac{\mathcal{B}_1(1)\mathcal{B}_{w-1}(n-1)}{\mathcal{B}_w(n)} \nonumber\\
    &= \frac{n}{w}
\end{align}
and the probability $A_{i+w} = 1$ is independent with $W_{i}$
\begin{align}
    P(A_{i+w} = 1) = \mathcal{B}_1(1) = p
\end{align}
The above two probabilities base on $Binomial$ assumption but not \emph{sum window}.

Note that in the transition process, $\sum_{n=1}^{w-1}A_{i+n}\cdot B_{i+n}$ is kept into the next state,
we name it with \emph{kept value}.
Abuse $\mathcal{N}_n(x)$ to represent the probability of $v = x$ for a random variable of $v \propto \mathcal{N}(n\mu, n\sigma^2)$.
When $W_i=x$, if $A_i = 1$,
the expectation of \emph{kept value} is:
\begin{align}
    \label{equ:exp_kept}
    & E_k(\sum_{n=1}^{w-1}A_{i+n}\cdot B_{i+n}|A_i = 1, W_i=x, S(W_{i}) = n) \nonumber\\
    =& \int_{-\infty}^{\infty} y\frac{\mathcal{N}_1(x-y)\mathcal{N}_{n - 1}(y)}{\mathcal{N}_n(x)}dy \nonumber\\
    =& \frac{n-1}{n}x
\end{align}
The result is intuitive since each of those $n$ records are i.i.d.
Otherwise if $A_i = 0$, it is obvious that \emph{kept value} is $x$.

The newly-incoming state $A_{i+w}\cdot B_{i+w}$ is independent with $W_{i}$ as well as the \emph{kept value}.
For \emph{sum window} the next state is the sum of $A_{i+w}\cdot B_{i+w}$ and \emph{kept value}.

Now use the similar method to make a brief analysis of the \emph{sum window} with \emph{Poisson} assumption.
Since $A_{i+w} \in [0, m]$, there are $(m + 1) * (n + 1)$ situations in the transition from $W_{i}$ to $W_{i + 1}$.
The detail of there situations is omitted but can be gotten easily.

Abuse $\mathcal{P}_w(x)$ to represent the probability of $v = x$ for a random variable of $v \propto \mathcal{P}(w, p)$.
$A_i \in [0, n]$ given $S(W_{i}) = n$, and the probability of each situation is
\begin{align}
    \label{equ:prob_poi_ai}
    P(A_i = a|S(W_{i}) = n) &= \frac{\mathcal{P}_1(a)\mathcal{P}_{w-1}(n-a)}{\mathcal{P}_w(n)} \nonumber\\
    &=\frac{n!(w-1)^{n-a}}{a!(n-a)!w^n}
\end{align}
where $a \in [0, n]$.
And the probability $A_{i+w} = b$ is independent with $W_{i}$ as
\begin{align}
    \label{equ:prob_poi_aiw}
    P(A_{i+w} = b) = \mathcal{P}_1(b) = \frac{p^b}{b!}e^{-p}
\end{align}
Above two probabilities base on $Poisson$ assumption but not \emph{sum window}.
The expectation of \emph{kept value} is correspondingly $\frac{n - a}{n}x$.

The expectation of \emph{kept value} for \emph{average window} with both assumptions can also be easily gotten with similar methods.
We will use them directly in the following sections without proof.

Overall, the transition kernels of those Markov Chains are very complex.
But we do not need to give an analytical expression of the transition kernel in this framework.

\subsection{Aggregation Estimation}

The aim of this Section is to estimate a possible area of mentioned three aggregators on the sliding window model.

Since both the sliding windows are modeled as Markov Chain with stationary distribution in GMM form,
without loss of generality, consider a stationary distribution $\pi = \sum_{n=0}^{mw} a_n \mathcal{N}(b_n\mu, c_n\sigma^2)$,
$\sum_{n=0}^{mw} a_n = 1$ and $a_n > 0$.
Let $\overline{b} = \sum_{n=0}^{mw} a_n b_n$ and the mean value of this distribution $\overline{\mu} = \overline{b}\mu$.

Denote $\{W_{i}^{n}\}, n \in [0, mw]$ as a subsequence which contains all the elements in $\{W_{i}\}_{1\leq i\leq \ell}$ where $S(W_{i}) = n$,
obviously the distribution of $\{W_{i}^{n}\}$ belongs to the $n$-th component of given GMM.
Note that $W_{i}^{n} \in [b_n\underline{\mathcal{C}}, b_n\overline{\mathcal{C}}]$ holds for both \emph{sum window} and \emph{average window},
we call it \emph{real bound}.

\subsubsection{Average Value}
\label{sec:avg}

Special cases for average aggregation that we are able to write the value directly are list as follows:
\begin{itemize}
    \item The average aggregation for \emph{sum window} with \emph{Always} assumption is $w\mu$.
    \item The average aggregation for \emph{average window} with \emph{Always} assumption is $\mu$.
\end{itemize}
Only \emph{Binomial} and \emph{Poisson} assumptions are taken into account.

Take \emph{average window} with \emph{Poisson} assumption as an example.
Define $\tau$, $\kappa$ and $\phi$ as follows:
\begin{equation}
    \tau=|\mu - \underline{\mathcal{C}}| \vee |\overline{\mathcal{C}} - \mu|
\end{equation}
\begin{equation}
\label{equ:kappa}
    \kappa = \sum_{n=0}^{mw} \sum_{a=0}^{n} \sum_{b=0}^{m} a_n \frac{n!(w-1)^{n-a}}{a!(n-a)!w^n} \frac{p^b}{b!}e^{-p} \frac{(n - a)}{n - a + b}
\end{equation}
\begin{equation}
\label{equ:phi}
    \phi = \sum_{n=0}^{mw} \sum_{a=0}^{n} \sum_{b=0}^{m} a_n \frac{n!(w-1)^{n-a}}{a!(n-a)!w^n} \frac{p^b}{b!}e^{-p} \frac{b}{n - a + b}
\end{equation}
and then let
\begin{equation}
    \label{equ:ph_inner_product}
    \lambda = \sqrt{\frac{\sum_{n=0}^{mw} a_n\kappa^2 \left( c_n\sigma^2 + (b_n - \frac{\overline{b} - \phi}{\kappa} )^2\mu^2 \right)}{\sum_{n=0}^{mw} a_n \left( c_n\sigma^2 + (b_n - \overline{b})^2\mu^2 \right)}}
\end{equation}

Now with a given p-value $\rho$ the following inequality
\begin{gather}
    \sum_{n=0}^{mw} a_nb_n\ell\mu - \sum_{n=0}^{mw} b_n\tau \sqrt{2\alpha(\lambda)a_n\ell \log{\frac{2}{1 - \rho}}} \notag \\
    \leq \sum_{i=1}^{\ell} W_{i} \leq \label{equ:w_bound} \\
    \sum_{n=0}^{mw} a_nb_n\ell\mu + \sum_{n=0}^{mw} b_n\tau \sqrt{2\alpha(\lambda)a_n\ell \log{\frac{2}{1 - \rho}}} \notag
\end{gather}
holds with probability at least $\rho$ and $\alpha : \lambda \mapsto (1 + \lambda)/(1 - \lambda)$.

The derivation is in the Supplement.
$\kappa$ and $\phi$ of other windows and assumptions can be derivated by similar method.

Recall that if $S(W_{i}) = 0$, this state will not be accounted as there is no record in this timestamp.
Thus we get
\begin{equation}
    AGG_{avg}(W(\mathcal{C}, w)) = \frac{\sum_{i=1}^{\ell} W_{i}}{\sum_{n=1}^{mw}a_n\ell}
\end{equation}

Note that if $a_n\ell$ is small enough to satisfy the inequality $a_n\ell < 2\alpha(\lambda) \log{\frac{2}{1 - \rho}}$,
the estimated bound in the $n$-th component will exceed the \emph{real bound} $[a_nb_n\ell \underline{\mathcal{C}}, a_nb_n\ell \overline{\mathcal{C}}]$.
So it is still necessary to use \emph{real bound} to bound the estimated bound in (\ref{equ:w_bound}) for each component.

\subsubsection{Max Value}
\label{sec:max}

A series of theoretical research focuses on the EVT of stochastic process or Markov Chain.
The asymptotic property with the extreme value of corresponding i.i.d sequence has been proved.
Let $\{X_i\}_{1\leq i\leq \ell}$ denote a stationary sequence with marginal distribution function $F$.
For large $\ell$ and $u$, it is typically the case that
\begin{equation}
    P(\max(X_1, \cdots, X_{\ell})\leq u) \thickapprox F(u)^{n\theta}
\end{equation}
where $\theta \in [0, 1]$ is a constant for the process known as the \emph{extremal index}.

Some works like~\cite{Smith1992The} and~\cite{Smith1994Estimating} focus on the calculation of \emph{extremal index} for general Markov Chain.
For some specific situations, $\theta$ has close-form solution.
But to our case, the calculation of $\theta$ depends on the iteration of integral and convolution,
it is too expensive to solve it.
Even though we do not apply the EVT for general Markov Chain,
these works are still useful for other types of sliding windows like max,
which can be modeled with Gumbel distribution.

Research on EVT of stationary Gaussian sequence is summarized in ~\cite{Leadbetter1983Extremes}.
As is proved in~\cite{Berman1964Limit},
for stationary Gaussian sequence $\{X_{i}\}$ with Gaussian distribution $\mathcal{N}(b_n\mu, c_n\sigma^2)$ as its stationary distribution,
define $\{\alpha_{i}\}$, $\{\beta_{i}\}$ and $\{\gamma_i\}$ as follows:
\begin{equation}
    \alpha_{i} = (2\log i)^{-\frac{1}{2}}
\end{equation}
\begin{equation}
    \beta_{i} = (2\log i)^{\frac{1}{2}} - \frac{1}{2}(2\log i)^{-\frac{1}{2}}(\log \log i + \log 4\pi)
\end{equation}
\begin{equation}
    \gamma_i = EX_{0}X_{i}
\end{equation}
below asymptotic holds if $\lim_{i \to \infty}\gamma_{i} \log i = 0$:
\begin{equation}
\label{equ:max_gau}
    P\left\{ \max(X_1, \cdots, X_i) \leq \mu + \sigma(\alpha_ix + \beta_i) \right\} \to e^{-e^{-x}}\\
\end{equation}

Note that all $\{W_{i}^{n}\}, n \in [0, mw]$ are stationary Gaussian sequences,
while the transition kernel of $\{W_{i}^{n}\}$ is much more complex than $\{W_{i}\}$.
The length for $\{W_{i}\}$ is asymptotic to be $\ell_n \to a_n\ell$.

In our case, $\gamma_i$, \ie~the corelationship of $\{W_{0}\}$ and $\{W_{i}\}$,
satisfies $\gamma_i = 0, \forall i > w$ since they are in fact independent.
Thus $\lim_{i \to \infty}\gamma_{i} \log i = 0$ holds for all $\{W_{i}^{n}\}$.

An asymptotic bound for the max value of $\{W_{i}^{n}\}$ can be estimated with 2 given p-values $\rho_l < \rho_r$ as:
\begin{gather}
    b_n\mu + \sqrt{c_n}\sigma(\alpha_{\ell_n} \log(1 / \log \frac{1}{\rho_l}) +\beta_{\ell_n}) \notag\\
    \leq \max(\{W_{i}^{n}\}) \leq \\
    b_n\mu + \sqrt{c_n}\sigma(\alpha_{\ell_n} \log(1 / \log \frac{1}{\rho_r}) + \beta_{\ell_n}) \notag
\end{gather}

The bound for the max value of $\{W_{i}\}$ can be gotten by:
\begin{equation}
    \max(\{W_i\}) = \max(\max(\{W_{i}^{1}\}), \cdots , \max(\{W_{i}^{mw}\}))
\end{equation}

And, of course, the max value for each $\{W_{i}^{n}\}$ can only be in $[b_n\underline{\mathcal{C}}, b_n\overline{\mathcal{C}}]$.

\subsubsection{Min Value}
\label{sec:min}

The min value can be converted into the estimation of max value as $-\max(\{-W_i\})$.
Simply use $-b_n$ instead of $b_n$ to get the stationary distribution of $\{-W_i\}$.
And the real bound of max value of $\{-W_i\}$ is $[-b_n\overline{\mathcal{C}}, -b_n\underline{\mathcal{C}}]$.

\subsection{Feature Selection}

In this framework we simply use Random Forest as the measure to rank and select features since it returns the feature importance in percentage and runs rapidly.
Another tree learning models, \eg~XGBoost~\cite{Chen2016XGBoost} and LightGBM~\cite{NIPS2017_6907}, are also competent.
In fact, this framework does not make many requests to the feature selection method.

The value bound, which can be treated as the distribution, of each feature in $T_f$ is estimated with the above steps.
Bagging~\cite{Breiman1996Bagging} strategy strongly enhances the generalization ability of learning algorithms by resampling from a given distribution.
As we finally get the estimated distribution but not explicit value,
similar thought is applied to improve the accuracy of this method.

A fake $T_f$ is sampled from the confidence bound and the importance by percent of features is stored.
Repeat this many times and finally output the average importance as the basis to select features.

\subsection{Complexity}
\label{sec:complexity}

The major time cost in this scenario is to scan table $T_a$.
The first brute-force method to generate real $T_f$ is to cut sub frames by time and then perform calculation iteration.
The iteration increases linearly with $\ell$.
The second method takes advantage of the sparsity of $T_a$,
$T_a$ is sorted by foreign key and time,
the iteration only operates a few nearest samples in the current scanning timestamp.
\cite{Arasu2004Resource} proposes some optimization on time and space complexity about the iteration,
the iteration only increases linearly with the count of records.

In our proposed method, the count of records and timestamps are parameterized into the close-form solution,
so the complexity of estimation is not related to them.
In all the estimation process of 3 aggregators, the calculation is separated into $mw$ components.
Suppose there are $f$ features in $T_a$ and $e$ entities in $T_e$.
The time complexity of this method in given period $w$ and corresponding $m$ is
\begin{equation}
    O = O(efwm)
\end{equation}

But note the calculation of (\ref{equ:kappa}) and (\ref{equ:phi}).
These two equations are both the sum of a tensor with $mw * \frac{mw}{2} * m$ elements,
while some intermediate results can be summed before multiplication to reduce memory cost.
The time complexity of this calculation is $O(efw^2m^3)$ while the space complexity is $O(efw^2m^2)$.
Under normal circumstances, this operation only takes a small part of time cost,
but in some extreme cases, \ie~$w$ and $m$ are very large, the time and memory consumption of this operation will be quite large.
In Supplement, we provide an alternative approach and corresponding analysis to solve this problem.

\section{Experiments}
\label{sec:exp}

Recall that we have not found an existing approach to do feature selection from original time series tables.
As there is no suitable baseline to compare with,
the experiment focus on the accuracy and speed comparison with the explicit generation of real $T_f$.

\subsection{Setting}

\begin{table}
\centering
\caption{Datasets in experiments.}
\label{tab:datasets}
\begin{tabular}{l|ccc}
\hline
Dataset &
Entities &
Records &
Classes
\\
\hline
Tianchi$^1$ & 20000 & 1830386 & 2 \\
PLAsTiCC$^2$ & 7848 & 1421705 & 14 \\
NFL$^3$ & 25043 & 1024164 & 751 \\
MotionSense$^4$ & 360 & 1412864 & 6 \\
Gas Sensors$^5$ & 100 & 928991 & 3 \\
\hline
\end{tabular}
\footnotesize{$^1$\url{https://tianchi.aliyun.com/competition/entrance/231607/information}}\\
\footnotesize{$^2$\url{https://www.kaggle.com/c/PLAsTiCC-2018}}\\
\footnotesize{$^3$\url{https://www.kaggle.com/zynicide/nfl-football-player-stats}}\\
\footnotesize{$^4$\url{https://www.kaggle.com/malekzadeh/motionsense-dataset}}\\
\footnotesize{$^5$\url{http://archive.ics.uci.edu/ml/datasets/Gas+sensors+for+home+activity+monitoring}}\\
\end{table}

We set up experiments with mentioned 6 sliding window aggregators on 5 open datasets listed in Table~\ref{tab:datasets}.
Parameters are set as $\rho = 0.9$, $\rho_l = 0.05$, $\rho_r = 0.95$.
Given some periods, we firstly wash the original datasets and drop some useless features,
and then estimate the importance of features by this framework,
finally, we generate real $T_f$ to get the actual feature importance.

\begin{table*}
\centering
\caption{Detail settings of datasets in experiments.}
\label{tab:setting}
\begin{tabular}{l|cccccc}
\hline
Dataset &
Resample Frequency &
Timestamps &
Assumption &
$m$ &
Periods &
Features
\\
\hline
Tianchi & 1 day & 352 & Poisson & 10 & $[7, 15, 30, 60]$ & 984 \\
PLAsTiCC & 1 day & 731 & Poisson & 3 & $[3 - 10]$ & 144 \\
NFL & 1 year & 68 & Poisson & 20 & $[3 - 10]$ & 1776 \\
MotionSense & 1 second & 16424 & Always & 1 & $[10, 20, ... , 100]$ & 720 \\
Gas Sensors & 1 second & 15393 & Always & 1 & $[10, 20, ... , 100]$ & 600 \\
\hline
\end{tabular}
\end{table*}

The period settings, assumptions, max appearance limit $m$ and the count of output features, timestamps are listed in Table~\ref{tab:setting}.
\textbf{Tianchi} dataset is special since it is an insurance dataset, so the period is set to follow human behavior period,
~\ie~week, half a month, one month and two months.

The implement of RandomForest in scikit-learn~\cite{scikit-learn} is used to do the final feature selection step.
Note that the curse of dimension~\cite{Bishop2006Pattern} exists in \textbf{MotionSense} and \textbf{Gas sensors} datasets,
so the count of trees is set as 5000 to return a stable feature importance list in these two datasets.
While for other datasets, the count of trees is 100 while the ensemble count is 10.

The platform is OSX laptop with 4C8T Intel Core i7@2.2 GHz CPU and 16GB RAM.

\subsection{Accuracy}

\begin{figure}
\centering
\includegraphics[width=\linewidth]{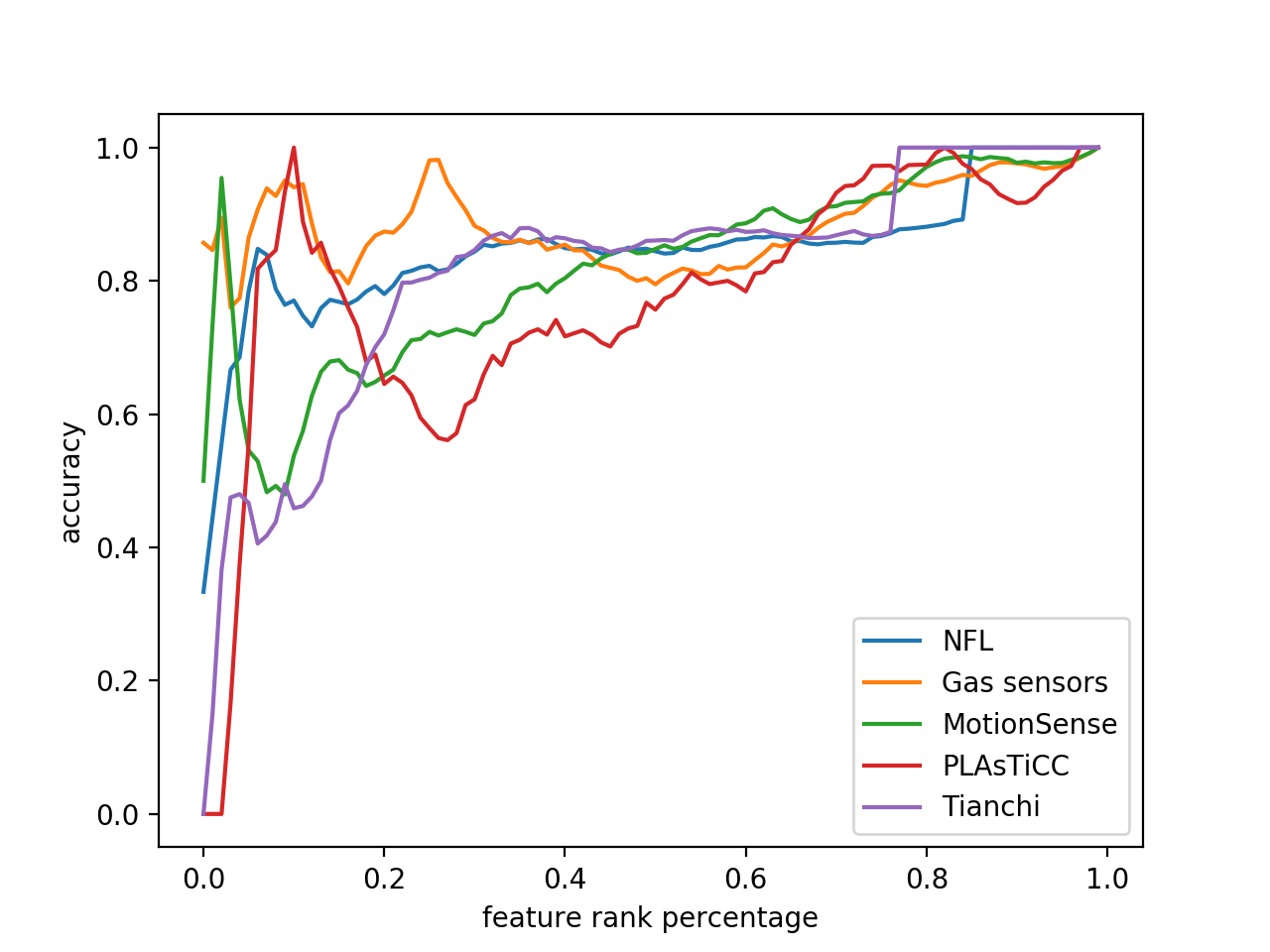}
\caption{Accuracy of output features selected by percentage. The accuracy is
calculated by the count of features whose actual rank and estimated rank are
both in the top n percent divided by the count of features whose actual rank
is in the corresponding place.}
\label{fig:line_rank}
\end{figure}

\begin{table}
\centering
\caption{Quartiles of relative error$^*$ between estimated and actual feature importance.}
\label{tab:error}
\begin{tabular}{l|ccc}
\hline
Dataset & 25\% & 50\% & 75\%
\\
\hline
Tianchi & -13\% & 12\% & 32\% \\
PLAsTiCC & -13\% & -5\% & 7\% \\
NFL & 2\% & 19\% & 41\% \\
MotionSense & -17\% & -5\% & 7\% \\
Gas Sensors & -8\% & 19\% & 46\% \\
\hline
\end{tabular}
\\
\footnotesize{$^*$The relative error is calculated by the difference between estimated
importance and actual importance divided by the actual importance.}\\
\end{table}

Fig.~\ref{fig:line_rank} shows the accuracy of features selected by importance rank percentage.
In \textbf{Tianchi}, \textbf{NFL} and \textbf{Gas sensors} datasets,
the accuracy increases quickly and achieves 0.8 in the top 20\% features.
Performance in \textbf{MotionSense} is to some extent worse,
but still achieves 0.7 in the top 25\% features.
In \textbf{PLAsTiCC} the result is worse but achieves more than 0.6 from the start.
As the count of features produced by \textbf{PLAsTiCC} is small,
the accuracy has obvious fluctuation.

Table~\ref{tab:error} lists the metric of relative error between estimated importance and actual importance by quartiles.
The estimated importance is very close to the actual importance in \textbf{PLAsTiCC}.
We think it is the count of features that reduce the accuracy in this dataset.

We set up experiments in datasets from different domains, with different features distribution,
class distribution, different sliding window periods, different assumptions,
and among which \textbf{MotionSense} and \textbf{Gas sensors} datasets are sensor data series,
which does not follow the basic assumption since the records are strongly related in the time dimension.
From the metric, we can see this framework achieves a high accuracy,
the relative error of estimated importance is small for most features.
With this accuracy, we can easily use this framework to estimate the importance and generate a bit more features
than expected count to cover the real important features.

\subsection{Speed}

\begin{table}
\centering
\caption{End-to-end Execution time cost.}
\label{tab:cost}
\begin{tabular}{l|ccc}
\hline
Dataset &
Ours &
DeepFeatureSynthesis &
Brute Force
\\
\hline
Tianchi & 1629s& 5586s & \\
PLAsTiCC & 26.6s & 4492s & 315s \\
NFL & 938s & 2410s & 1590s \\
MotionSense & 3.9s & 23h & 59s \\
Gas Sensors & 1.9s & 9h & 24s \\
\hline
\end{tabular}
\end{table}

We implement the two brute-force methods mentioned in Section~\ref{sec:complexity}.
Method 1 is the \emph{cutoff\_time} feature of \textbf{FeatureTools}\footnote{\url{https://www.featuretools.com/}},
which is the implement of Deep Feature Synthesis mentioned in Section~\ref{sec:related}.
Method 2 is implemented with \textbf{Pandas}\footnote{\url{https://pandas.pydata.org/}}.
And our method is implemented with \textbf{Numpy}\footnote{\url{https://numpy.org/}} to keep the same iteration efficiency.
The count of trees to evaluate feature importance is 100 for all, while other settings keep the same.

Table~\ref{tab:cost} list the time cost of end-to-end evaluation of three methods.
Since the window related functions of \textbf{Pandas} do not support duplicated time index,
the time cost data in \textbf{Tianchi} dataset is omitted.

The performance obeys the analysis in Section~\ref{sec:complexity},
with the increase of count of timestamps and records,
the advance of our proposed method grows.
The major part of time consumption of the estimation process is concentrated in sliding windows with large $w$ and $m$.
If we remove the period setting $w=60$, the end-to-end time cost of our method in \textbf{Tianchi} dataset reduces to 860 seconds.

For datasets with small $e$, $f$, $w$ and $m$, the major time cost is the feature selection.
The estimation process of sliding window aggregators only consumes 0.15 seconds in \textbf{Gas sensors},
while the calculation process of feature importance takes 1.3 seconds.
There is still some space for optimization about feature importance algorithm.

\subsection{Factor Analysis}

\begin{figure}
\centering
\includegraphics[width=\linewidth]{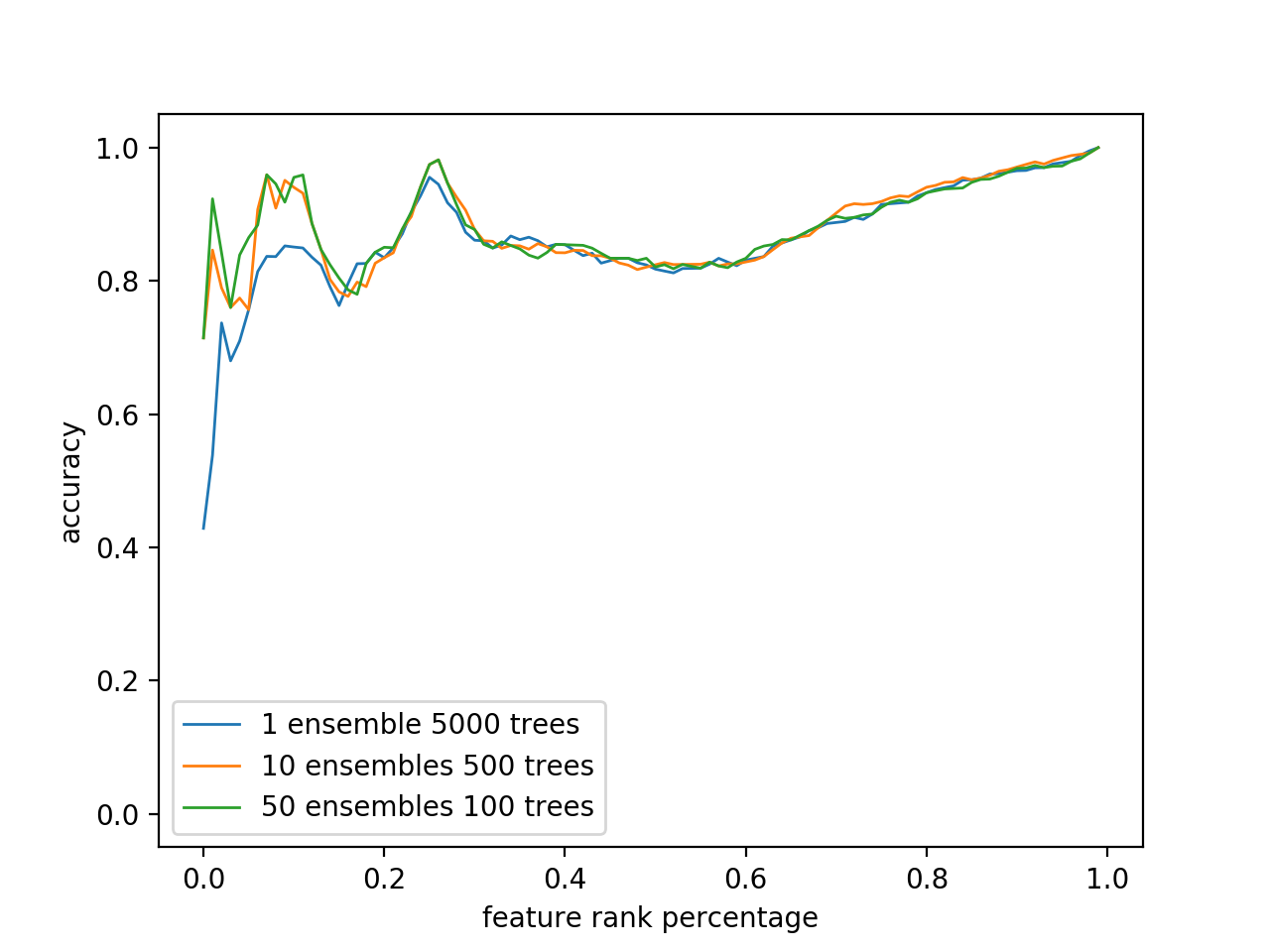}
\caption{Accuracy of different ensemble settings in \textbf{Gas sensors} dataset.}
\label{fig:ensemble_comp}
\end{figure}

We set up a control experiment to compare the effect of ensemble.
Due to lack of space, Fig.~\ref{fig:ensemble_comp} only provide the comparison in \textbf{Gas sensors} dataset,
and appearance is similar in other datasets.
The total evaluation is same to be 5000 trees, while the ensemble count and the count of trees in each ensemble is different.

Obviously ensemble strategy enhances the accuracy to some extent,
and it follows the law of diminishing marginal utility.
It is better to do more ensembles when total evaluation is certain.
Compared with the performance of the original experiment (10 ensembles and 5000 trees),
the performance when the ensemble count is 10 or 50 is similar, we think it reaches the supermum.

In this framework, when dataset and periods are certain,
the only variables except for ensemble count are $\rho$, $\rho_l$ and $\rho_r$.
From the behavior we think the supermum is certain with certain $\rho$, $\rho_l$ and $\rho_r$.
Unfortunately, there is not enough theory to describe the relationship between the supermum and those variables.
And the experiments with different $\rho$, $\rho_l$ and $\rho_r$ have not reflect some intuitive rules.
Thus we fail to propose a simple and efficient strategy to determine the value of those variables to optimize the supermum.
Maybe research about the influence on the separability of features by disturbance can help this topic.

\section{Conclusions and Future Work}
\label{sec:conclusion}

In this paper, we propose a framework to fill the gap of the end-to-end automatic sliding window aggregate feature selection for time series.
This framework encodes the actions of entities into distributions and selects features from distributions directly.
Even though this framework depends on some distribution assumptions,
those assumptions are not strong and they greatly simplify the theoretical derivation.
The employment of ensemble strategy guarantees the accuracy of feature importance estimation.
And the complexity of this framework is low in most circumstances.
The type of sliding windows and aggregators are easy to extend with existing theories about stochastic process.
From the experiments, the empirical performance of this framework is good enough to be applied in real-world scenarios.
With its low complexity, it is able to process a mass of period settings and features.

But there are still some challenges call for further research.
The mistake bound is not guaranteed in theory, and how do the factors and datasets influence the accuracy is also unknown.
\section{Supplement}

\subsection{Derivation of Average Value Estimation}
\label{sec:hoeffding_derivation}

Fan~\etal~\cite{2018arXiv180200211F} have proved following Hoeffding's lemma:
\begin{equation}
\label{equ:avg}
    \mathbb{P}_{\pi}\left( \left| \sum_{i=1}^{n} f_i(X_i) - \sum_{i=1}^{n}\pi(f_i) \right| > \epsilon \right) \leq 2e^{-\frac{\alpha(\lambda)^{-1}\epsilon^2}{2 \sum_{i=1}^{n} (u_i-l_i)^2/4}}
\end{equation}
holds for general Markov Chain $\{X_i\}_{1\leq i\leq n}$ with invariant measure $\pi$ and spectral gap $1 - \lambda > 0$,
$f_i : \mathbb{R} \mapsto [u_i,l_i]$ is a time-dependent bounded function,
$\alpha : \lambda \mapsto (1 + \lambda)/(1 - \lambda)$ and
\begin{equation}
\label{equ:pif}
    \pi(f) = \int f(x)\pi(dx)
\end{equation}
(\ref{equ:pif}) can be viewed as the expectation of $f(x)$, $x\propto \pi$.
Also it is is the sharpest bound it can be in theory for general Markov Chain

In our case, if the bound should be symmetry by the mean value of a Gaussian distribution to make $\pi(f)$ integrable.
Let $\tau=|\mu - \underline{\mathcal{C}}| \vee |\overline{\mathcal{C}} - \mu|$,
and design time-dependent map function $f_n$ for the $n$-th component of given GMM as:
\begin{equation}
    f_n(x) = \left\{
        \begin{array}{lcl}
            b_n\mu-b_n\tau & x \leq b_n\mu-b_n\tau\\
            x              & b_n\mu-b_n\tau < x < b_n\mu+b_n\tau\\
            b_n\mu+b_n\tau & x \geq b_n\mu+b_n\tau
        \end{array}
    \right.
\end{equation}
Thus
\begin{align}
    \pi(f_n) = \int f_n(x)\mathcal{N}(b_n\mu, c_n\sigma^2)dx = b_n\mu
\end{align}
\begin{align}
    \label{equ:sum_pif}
    \sum_{i=1}^{\ell}\pi(f_i) = \sum_{n=0}^{mw} a_n\ell b_n\mu = \overline{b}\ell\mu
\end{align}
\begin{align}
    \label{equ:mapbound}
    u_n - l_n = 2 b_n\tau
\end{align}
And $f_n(W_i) = W_i$ holds in given samples from $\mathcal{C}$
since the actual observations in the $n$-th Gaussian component $W_{i}^{n} \in [b_n\mu-b_n\tau,b_n\mu+b_n\tau]$ for both sliding windows.

Transition kernel (from $x$ to $y$) of our Markov Chain cannot be transformed into a León-Perron operator
(see Definition 3.4 in~\cite{2018arXiv180200211F}).
So the spectral gap $\lambda$ can only be gotten by solving the equation
(see Definition 3.6 in~\cite{2018arXiv180200211F}) as follows:
\begin{align}
    \lambda &:= \sup\left\{ \frac{\lVert \int h(y)P(x, dy) \rVert_\pi}{\lVert h(x) \rVert_\pi} \right\}
\end{align}
where $h$ is any real-valued $\mathbb{B}$-measurable function $h : \mathbb{R} \mapsto \mathbb{R}$, $\pi(h)=0, h\neq 0$.
And the inner product of a function $h$ on $\pi$ is defined as 
\begin{equation}
    \label{eqa:inner_product}
    \lVert h(x) \rVert_\pi = \sqrt{\int h(x)h(x)\pi(dx)}
\end{equation}

The explicit value of $\lambda$ can be calculated by following steps.
Use map function $h : x \mapsto x - \overline{\mu}$ to satisfy $\pi(h)=0, h\neq 0$.
Firstly we calculate a useful integral
\begin{align}
    \label{equ:integral}
    &\int_{-\infty}^{\infty} (x - k\mu)(x - j\mu) \mathcal{N}(b_n\mu, c_n\sigma^2) dx \nonumber\\
    = & c_n\sigma^2 + (b_n - k)(b_n - j)\mu^2
\end{align}
holds for arbitrary $k$ and $j$.
And it is easy to use (\ref{equ:integral}) to calculate $\lVert h(x) \rVert_\pi$:
\begin{equation}
    \label{equ:hx_inner_product}
    \lVert h(x) \rVert_\pi = \sqrt{\sum_{n=0}^{mw} a_n \left( c_n\sigma^2 + (b_n - \overline{b})^2\mu^2 \right)}
\end{equation}
and $\int h(y)P(x, dy)$:
\begin{align}
    \label{equ:phx}
    \int h(y)P(x, dy) & = \int yP(x, dy) - \overline{\mu}\int P(x, dy) \nonumber\\
    & = \int yP(x, dy) - \overline{\mu}
\end{align}
The first term of (\ref{equ:phx}) is the expectation of $W_{i+1}$ given $W_{i} = x$.
As analyzed before, this expectation can be separated into discrete situations by $A_i$ and $A_{i+w}$.
It is composed with the expectation of \emph{kept value} and the expectation of incoming timestamp in each situation since they are independent.

\begin{table}
\centering
\caption{Expectation of $W_{i+1}$ given $W_{i} = x$ when $S(W_i) = n$ with \emph{Binomial} assumption.}
\label{tab:phx}
\begin{tabular}{l|c|c|c}
\hline
Window & Situation & Coeffient & Expectation \\
\hline
sum & 1 & $\frac{n}{w}p$ & $\frac{(n - 1)}{n}x + \mu$ \\
sum & 2 & $\frac{n}{w}(1 - p)$ & $\frac{(n - 1)}{n}x$ \\
sum & 3 & $(1 - \frac{n}{w})p$ & $x + \mu$ \\
sum & 4 & $(1 - \frac{n}{w})(1 - p)$ & $x$ \\
\hline
avg & 1 & $\frac{n}{w}p$ & $\frac{(n - 1)x + \mu}{n}$ \\
avg & 2 & $\frac{n}{w}(1 - p)$ & $x$ \\
avg & 3 & $(1 - \frac{n}{w})p$ & $\frac{nx + \mu}{n + 1}$ \\
avg & 4 & $(1 - \frac{n}{w})(1 - p)$ & $x$ \\
\hline
\end{tabular}
\end{table}

The transition kernel is formed with four parts for \emph{Binomial} assumption,
Table \ref{tab:phx} lists the expectation of $W_{i+1}$ given $W_{i} = x$ when $S(W_i) = n$,
while the Coefficient column is the probability of the corresponding situation.
And $\int h(y)P(x, dy)$ is the weighted sum of those four situations.

For \emph{Poisson} assumption, take \emph{average window} as an example.
Consider $A_i = a$ and $A_{i+w} = b$ with arbitary $a \in [0, n]$ and $b \in [0, m]$ when $S(W_i) = n$,
the expectation of $W_{i+1}$ given $A_i = a$, $A_{i+w} = b$ and $W_{i} = x$ is
\begin{align}
    &E(W_{i+1} | A_i = a, A_{i+w} = b, W_{i} = x, S(W_i) = n) \nonumber\\
    =& \frac{(n - a)}{n - a + b}x + \frac{b}{n - a + b}\mu
\end{align}
The corresponding probability of this situation is described in transition kernel section.

Define $\kappa$ and $\phi$ to represent the final coefficient of $x$ and $\mu$ in (\ref{equ:phx}) as
\begin{equation}
    \int h(y)P(x, dy) = \kappa x - (\overline{b} - \phi)\mu
\end{equation}
For \emph{average window} with \emph{Poisson} assumption
\begin{align}
    \kappa = \sum_{n=0}^{mw} \sum_{a=0}^{n} \sum_{b=0}^{m} a_n \frac{n!(w-1)^{n-a}}{a!(n-a)!w^n} \frac{p^b}{b!}e^{-p} \frac{(n - a)}{n - a + b}
\end{align}
\begin{align}
    \phi = \sum_{n=0}^{mw} \sum_{a=0}^{n} \sum_{b=0}^{m} a_n \frac{n!(w-1)^{n-a}}{a!(n-a)!w^n} \frac{p^b}{b!}e^{-p} \frac{b}{n - a + b}
\end{align}
Other situations are similar.
In fact, \emph{Binomial} assumption can be treated as special case with $a \in [0, 1]$ and $b \in [0, 1]$.

The inner product of $\int h(y)P(x, dy)$ can now be gotten:
\begin{align}
    \label{equ:ph_inner_product}
    &\lVert \int h(y)P(x, dy) \rVert_\pi \nonumber\\
    =& \sqrt{\sum_{n=0}^{mw} a_n\kappa^2 \left( c_n\sigma^2 + (b_n - \frac{\overline{b} - \phi}{\kappa} )^2\mu^2 \right)}
\end{align}
And then $\lambda$ is gotten by (\ref{equ:ph_inner_product}) and (\ref{equ:hx_inner_product}).

Rewrite (\ref{equ:avg}) with given p-value $\rho$, (\ref{equ:mapbound}) and (\ref{equ:sum_pif}), inequality
\begin{gather}
    \sum_{n=0}^{mw} a_nb_n\ell\mu - \sum_{n=0}^{mw} b_n\tau \sqrt{2\alpha(\lambda)a_n\ell \log{\frac{2}{1 - \rho}}} \notag \\
    \leq \sum_{i=1}^{\ell} W_{i} \leq \\
    \sum_{n=0}^{mw} a_nb_n\ell\mu + \sum_{n=0}^{mw} b_n\tau \sqrt{2\alpha(\lambda)a_n\ell \log{\frac{2}{1 - \rho}}} \notag
\end{gather}
holds with probability at least $\rho$.

\subsection{Alternative Approach of $\lambda$}


\begin{figure}
\centering
\includegraphics[width=\linewidth]{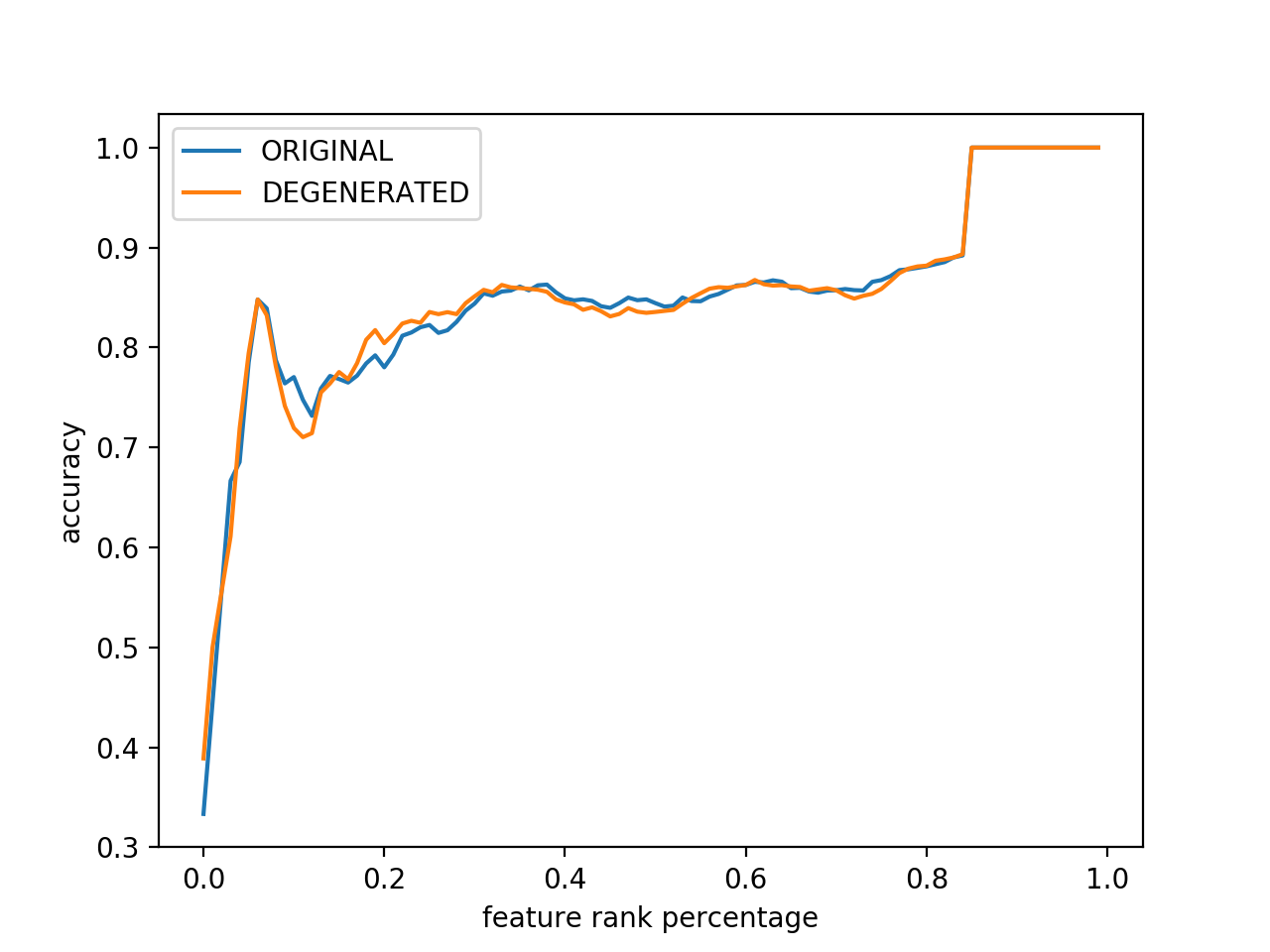}
\caption{Accuracy comparison between original and degenerated method in \textbf{NFL}.}
\label{fig:degenerated}
\end{figure}

The fourth situation, \ie~$A_i = 0$ and $A_{i+w} = 0$, is special since the state will not transfer.
Let $\mathbb{B}$ be the Borel $\sigma$-algebra over real space $\mathbb{R}$,
then we can write the transition kernel directly for this situation as:
\begin{align}
    P_4(W_i,X) &= P(W_{i+1} \in X|W_i), \forall X \in \mathbb{B}, \forall i \geq 0\nonumber\\
    &=\mathbb{I}(W_i \in X), \forall W_i \in \mathbb{R}, \forall X \in \mathbb{B}
\end{align}
where $\mathbb{I}$ is the indicator operator, you can find further description in~\cite{2018arXiv180200211F}.

A degenerated method is to use the probability of the fourth situation.
For \emph{Binomial} assumption, the spectral gap satisfies
\begin{equation}
    \lambda > \underline{\lambda} = \sum_{n=0}^{N} (1 - \frac{n}{N})(1 - p)
\end{equation}
(see Definition 3.4 in~\cite{2018arXiv180200211F}).
As the coefficient function $\alpha^{-1}$ of (\ref{equ:avg}) is strictly decreasing with $\lambda$,
a \emph{tighter} bound is gotten by replacing $\lambda$ with $\underline{\lambda}$.
Since the product of big matrices is omitted,
this approach eliminates the high-order component of the time and space complexity.

But in general, the closer $\underline{\lambda}$ is to 0, the closer $\lambda$ is to 1 under \emph{Binomial} assumption and \emph{Poisson} assumption.
Which means $\lambda$ can be many times larger than $\underline{\lambda}$ when $p$ is large,
therefore the estimated bound will be many times tighter than it should be.
So the influence of this approach can not be estimated simply.

We test the performance in three datasets whose assumption is \emph{Poisson}.
P-value $\rho = 0.999$ to weaken the influence.
The metric of $\underline{\lambda}$ is similar to the original $\lambda$,
while time cost decreases by a some proportion.
We show the result in \textbf{NFL} in Fig.~\ref{fig:degenerated}.

\bibliographystyle{IEEEtran}
\bibliography{IEEEabrv,total}

\begin{thebibliography}{10}
\providecommand{\url}[1]{#1}
\csname url@samestyle\endcsname
\providecommand{\newblock}{\relax}
\providecommand{\bibinfo}[2]{#2}
\providecommand{\BIBentrySTDinterwordspacing}{\spaceskip=0pt\relax}
\providecommand{\BIBentryALTinterwordstretchfactor}{4}
\providecommand{\BIBentryALTinterwordspacing}{\spaceskip=\fontdimen2\font plus
\BIBentryALTinterwordstretchfactor\fontdimen3\font minus
  \fontdimen4\font\relax}
\providecommand{\BIBforeignlanguage}[2]{{%
\expandafter\ifx\csname l@#1\endcsname\relax
\typeout{** WARNING: IEEEtran.bst: No hyphenation pattern has been}%
\typeout{** loaded for the language `#1'. Using the pattern for}%
\typeout{** the default language instead.}%
\else
\language=\csname l@#1\endcsname
\fi
#2}}
\providecommand{\BIBdecl}{\relax}
\BIBdecl

\bibitem{Lam2017One}
H.~T. Lam, J.~M. Thiebaut, M.~Sinn, C.~Bei, T.~Mai, and O.~Alkan, ``One button
  machine for automating feature engineering in relational databases,'' 2017.

\bibitem{Kanter2015Deep}
J.~M. Kanter and K.~Veeramachaneni, ``Deep feature synthesis: Towards
  automating data science endeavors,'' in \emph{IEEE International Conference
  on Data Science \& Advanced Analytics}, 2015.

\bibitem{Tang2014Feature}
J.~Tang, S.~Alelyani, and H.~Liu, ``Feature selection for classification: A
  review,'' \emph{Documentación Administrativa}, pp. 313--334, 2014.

\bibitem{Molina2003Feature}
L.~C. Molina, L.~Belanche, and Àngela Nebot, ``Feature selection algorithms: A
  survey and experimental evaluation,'' in \emph{IEEE International Conference
  on Data Mining}, 2003.

\bibitem{Li:2019:AUF:3292500.3330856}
\BIBentryALTinterwordspacing
J.~Li, R.~Guo, C.~Liu, and H.~Liu, ``Adaptive unsupervised feature selection on
  attributed networks,'' in \emph{Proceedings of the 25th ACM SIGKDD
  International Conference on Knowledge Discovery \& Data Mining}, ser. KDD
  '19.\hskip 1em plus 0.5em minus 0.4em\relax New York, NY, USA: ACM, 2019, pp.
  92--100. [Online]. Available:
  \url{http://doi.acm.org/10.1145/3292500.3330856}
\BIBentrySTDinterwordspacing

\bibitem{2017A}
H.~U. Xuegang, P.~Zhou, L.~I. Peipei, J.~Wang, and W.~U. Xindong, ``A survey on
  online feature selection with streaming features,'' \emph{Frontiers of
  Computer ence}, vol.~12, no.~3, 2017.

\bibitem{2018Online}
D.~You, X.~Wu, L.~Shen, Z.~Chen, and S.~Deng, ``Online feature selection for
  streaming features with high redundancy using sliding-window sampling,'' in
  \emph{2018 IEEE International Conference on Big Knowledge (ICBK)}, 2018.

\bibitem{Hellerstein1997Online}
J.~M. Hellerstein, P.~J. Haas, and H.~J. Wang, ``Online aggregation,''
  \emph{Acm Sigmod Record}, vol.~26, no.~2, pp. 171--182, 1997.

\bibitem{2012arXiv1201.0559C}
K.-M. {Chung}, H.~{Lam}, Z.~{Liu}, and M.~{Mitzenmacher}, ``{Chernoff-Hoeffding
  Bounds for Markov Chains: Generalized and Simplified},'' \emph{arXiv
  e-prints}, p. arXiv:1201.0559, Jan 2012.

\bibitem{2018arXiv180611519R}
S.~{Rao}, ``{A Hoeffding inequality for Markov chains},'' \emph{arXiv
  e-prints}, p. arXiv:1806.11519, Jun 2018.

\bibitem{2018arXiv180200211F}
J.~{Fan}, B.~{Jiang}, and Q.~{Sun}, ``{Hoeffding's lemma for Markov Chains and
  its applications to statistical learning},'' \emph{arXiv e-prints}, p.
  arXiv:1802.00211, Feb 2018.

\bibitem{O1987Extreme}
G.~L. O'Brien, ``Extreme values for stationary and markov sequences,''
  \emph{Annals of Probability}, vol.~15, no.~1, pp. 281--291, 1987.

\bibitem{Rootz1988Maxima}
H.~Rootzén, ``Maxima and exceedances of stationary markov chains,''
  \emph{Advances in Applied Probability}, vol.~20, no.~2, pp. 371--390, 1988.

\bibitem{Leadbetter1988Extremal}
M.~R. Leadbetter and H.~Rootzén, ``Extremal theory for stochastic processes,''
  \emph{Annals of Probability}, vol.~16, no.~2, pp. 431--478, 1988.

\bibitem{Perfekt1994Extremal}
R.~Perfekt, ``Extremal behaviour of stationary markov chains with
  applications,'' \emph{Annals of Applied Probability}, vol.~4, no.~2, pp.
  529--548, 1994.

\bibitem{Perfekt1997Extreme}
------, ``Extreme value theory for a class of markov chains with values in
  rd,'' \emph{Advances in Applied Probability}, vol.~29, no.~1, pp. 138--164,
  1997.

\bibitem{Gikhman2006A}
I.~I. Gikhman, ``A limit theorem for the number of maxima in the sequence of
  random variables in a markov chain,'' \emph{Theory of Probability \& Its
  Applications}, vol.~3, no.~2, pp. 166--172, 2006.

\bibitem{Kevin2012Mlapp}
K.~P. Murphy, \emph{Machine Learning, a Probabilistic Perspective}.\hskip 1em
  plus 0.5em minus 0.4em\relax The MIT Press, 2012, pp. 337--339.

\bibitem{Smith1992The}
R.~L. Smith, ``The extremal index for a markov chain,'' \emph{Journal of
  Applied Probability}, vol.~29, no.~1, pp. 37--45, 1992.

\bibitem{Smith1994Estimating}
R.~L. Smith and I.~Weissman, ``Estimating the extremal index,'' \emph{Journal
  of the Royal Statistical Society}, vol.~56, no.~3, pp. 515--528, 1994.

\bibitem{Leadbetter1983Extremes}
M.~R. Leadbetter, G.~Lindgren, and H.~Rootzén, \emph{Extremes and related
  properties of random sequences and processes}.\hskip 1em plus 0.5em minus
  0.4em\relax Springer-Verlag, 1983, ch.~4.

\bibitem{Berman1964Limit}
S.~M. Berman, ``Limit theorems for the maximum term in stationary sequences,''
  \emph{Annals of Mathematical Statistics}, vol.~35, no.~2, pp. 502--516, 1964.

\bibitem{Chen2016XGBoost}
T.~Chen and C.~Guestrin, ``Xgboost: A scalable tree boosting system,'' in
  \emph{Acm Sigkdd International Conference on Knowledge Discovery \& Data
  Mining}, 2016.

\bibitem{NIPS2017_6907}
G.~Ke, Q.~Meng, T.~Finley, T.~Wang, W.~Chen, W.~Ma, Q.~Ye, and T.-Y. Liu,
  ``Lightgbm: A highly efficient gradient boosting decision tree,'' in
  \emph{Advances in Neural Information Processing Systems 30}, I.~Guyon, U.~V.
  Luxburg, S.~Bengio, H.~Wallach, R.~Fergus, S.~Vishwanathan, and R.~Garnett,
  Eds.\hskip 1em plus 0.5em minus 0.4em\relax Curran Associates, Inc., 2017,
  pp. 3146--3154.

\bibitem{Breiman1996Bagging}
L.~Breiman, ``Bagging predictors,'' \emph{Machine Learning}, vol.~24, no.~2,
  pp. 123--140, 1996.

\bibitem{Arasu2004Resource}
A.~Arasu and J.~Widom, ``Resource sharing in continuous sliding-window
  aggregates,'' in \emph{Thirtieth International Conference on Very Large Data
  Bases}, 2004.

\bibitem{scikit-learn}
F.~Pedregosa, G.~Varoquaux, A.~Gramfort, V.~Michel, B.~Thirion, O.~Grisel,
  M.~Blondel, P.~Prettenhofer, R.~Weiss, V.~Dubourg, J.~Vanderplas, A.~Passos,
  D.~Cournapeau, M.~Brucher, M.~Perrot, and E.~Duchesnay, ``Scikit-learn:
  Machine learning in {P}ython,'' \emph{Journal of Machine Learning Research},
  vol.~12, pp. 2825--2830, 2011.

\bibitem{Bishop2006Pattern}
C.~M. Bishop, \emph{Pattern Recognition and Machine Learning}.\hskip 1em plus
  0.5em minus 0.4em\relax Springer Science+Business Media, 2006.

\end{thebibliography}

\end{document}